\documentclass[10pt,twocolumn,letterpaper]{article}

\usepackage{cvpr}
\usepackage{times}
\usepackage{epsfig}
\usepackage{graphicx}
\usepackage{amsmath}
\usepackage{amssymb}
\usepackage{subfig}
\usepackage{float}
\usepackage{math}
\usepackage{enumitem}
\usepackage{xcolor}
\usepackage{multirow}
\usepackage{booktabs}
\def\xtarget{x^\tau}
\def\xsourcei{x^i}
\def\xgen{\hat{x}}
\def\posetarget{p^\tau}
\usepackage[pagebackref=true,breaklinks=true,letterpaper=true,colorlinks,bookmarks=false]{hyperref}

\cvprfinalcopy % *** Uncomment this line for the final submission
\begin{document}

%%%%%%%%% TITLE
\title{Attention-based Fusion for Multi-source Human Image Generation}

\author{St\'{e}phane~Lathuili\`{e}re$^1$,
  ~Enver~Sangineto$^1$,
          ~Aliaksandr~Siarohin$^1$ and
        ~Nicu~Sebe$^{1,2}$\\
 	$^1$DISI, University of Trento, via Sommarive 14, Povo (TN), Italy\\
	$^2$Huawei Technologies Ireland, Dublin, Ireland\\
{\tt\small \{stephane.lathuiliere, enver.sangineto, aliaksandr.siarohin, niculae.sebe\}@unitn.it}\\
}

\maketitle
%\thispagestyle{empty}

%%%%%%%%% ABSTRACT
\begin{abstract}
We present a generalization of the person-image generation task, in which a human image is generated conditioned on a target pose and  {\em a set} $\Xvect$ of source appearance images. In this way, we can exploit  multiple, possibly complementary images of the same person which are usually available at training and at testing time. The solution we propose is mainly based on a local attention mechanism which selects relevant information from different source image regions, avoiding the necessity to build specific generators for each specific cardinality of $\Xvect$. The empirical evaluation of our method  shows the practical interest of addressing the person-image generation problem in a multi-source setting.

\end{abstract}

%%%%%%%%% BODY TEXT

\vspace{-0.5cm}
\section{Introduction}
\label{Introduction}

The person image generation task, as proposed by Ma et al. \cite{ma2017pose},  consists in generating  ``person images in arbitrary poses, based on an image of that person and a novel pose''.
This task has recently attracted  a lot of interest in the community 
because of different potential applications, such as 
computer-graphics based manipulations \cite{walker2017pose}
 or data augmentation for training
    person re-identification   \cite{Zheng_2017_ICCV,liu2018pose}
   or  human pose estimation \cite{Cao} systems. 
Previous work on this field
\cite{ma2017pose,LassnerPG17,ZhaoWCLF17,siarohin2018deformable,balakrishnansynthesizing,si2018multistage}
assume that the generation task is conditioned on two variables: the appearance image of a person (we call this variable the {\em source} image) and a {\em target} pose, automatically extracted from a different image of the same person using a 
Human Pose Estimator (HPE).

Using person-specific abundant data
 the quality of
 the generated images  can be potentially improved. For instance, a training dataset specific to each target person can be recorded \cite{chan2018dance}. Another solution is to build a full-3D model of the  target person \cite{liu2018neural}.  However, these approaches lack of flexibility and
need an
expensive data-collection. 
\begin{figure}[t!]
\centering
\includegraphics[width=0.80\columnwidth]{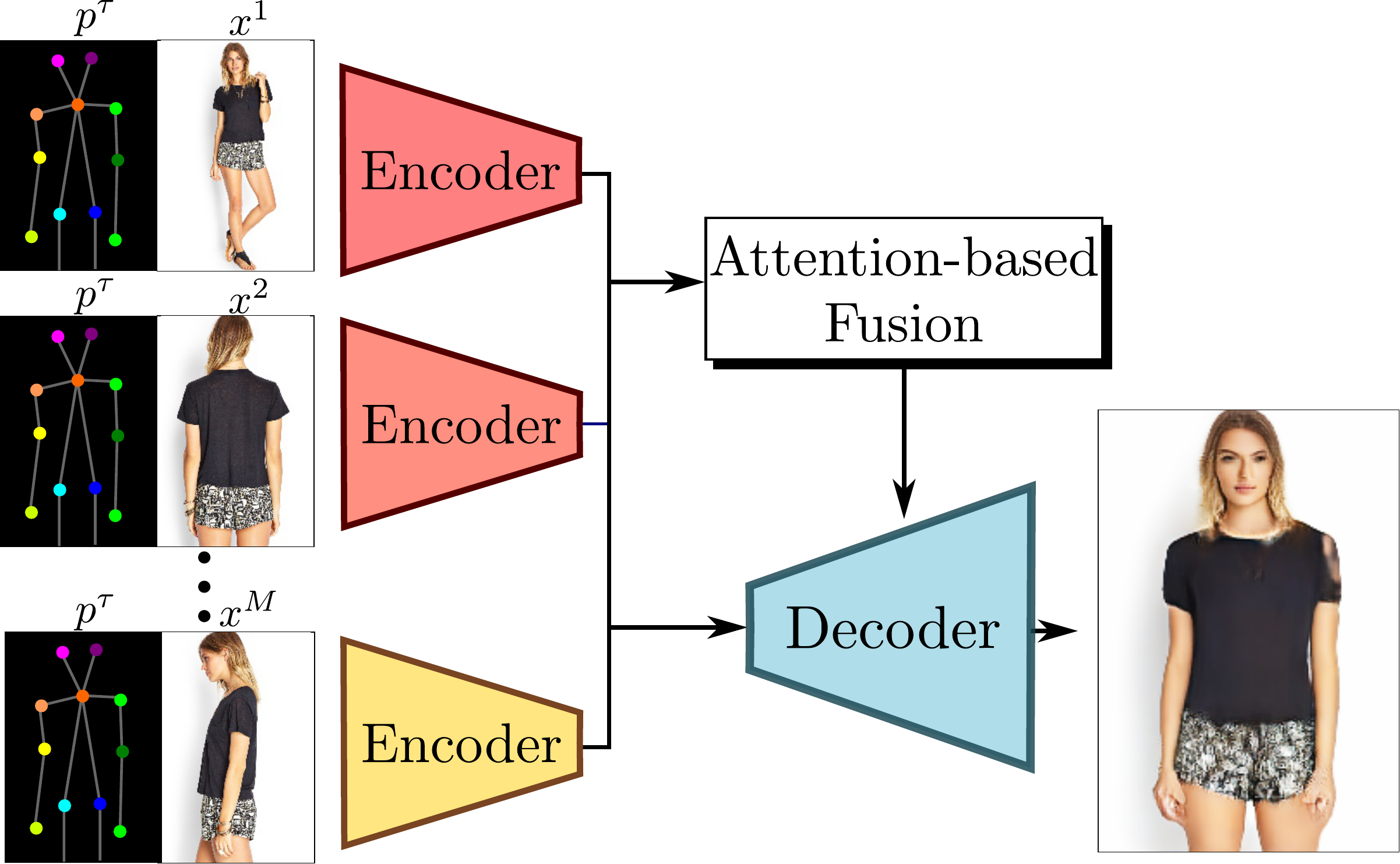}\label{fig:teaser}
\caption{Multi-source Human Image Generation: an image of a person in a novel pose is generated from {\em a set} of images of the same person.}
\vspace{-0.6cm}
\label{fig:teaser}
\end{figure}

In this work we propose a different direction which relies on  a few, variable number of source images (e.g., from 2 to 10). We call the corresponding task {\em multi-source human image generation}.
As far as we know, no previous work has investigated this direction yet.
The reason for which we believe this generalization of the person-image generation task is interesting is that multiple source images, when available, can provide  richer appearance information. This data redundancy   can possibly be exploited by the generator in order to compensate for partial occlusions, self-occlusions or noise in the source images.
More formally, we define our multi-source human image generation task as follows.
We assume that a set
 of $M$ ($M \geq 1$) source images $\Xvect=\{\xsourcei\}_{i=1..M}$ 
is given and that these images depict the same person with the same 
overall appearance (e.g., the same clothes, haircut, etc.).  Besides, a unique
 target body pose $\posetarget$ is provided, typically extracted from a target image not contained in 
 $\Xvect$.
The multi-source human image generation task consists in generating a new image $\xgen$ with an appearance  similar to the general appearance 
pattern 
represented in $\Xvect$  but in the pose $\posetarget$ (see Fig.~\ref{fig:teaser}).
Note that $M$ is not a-priori fixed, and we believe this task characteristics are important for practical applications, in which the same dataset can contain multiple-source images of the same person but with  unknown and variable cardinalities. 
%In addition, from a user perspective, providing several images of a person instead of only a single one %is not much more practically constraining.    
%% For instance, in person re-identification datasets, a given person identity is typically represented by groups of images with a variable cardinality, hence the generation process should be able to exploit all the available, possibly redundant, source images for each different person.

Most 
of previous methods on single-source human image generation 
\cite{siarohin2018deformable,LassnerPG17,ma2017pose,walker2017pose,ZhaoWCLF17,esser2018variational,si2018multistage,liu2018pose}
are based on variants of the U-Net architecture generator proposed by Isola et al. \cite{pix2pix2016}. A common, general idea in these methods is that the conditioning information (e.g., the source image and/or the target pose) is  
transformed into the desired synthetic image using the U-Net {\em skip connections}, which shuttle information between those layers in the encoder and in the decoder having a
corresponding resolution (see Sec.~\ref{method}).
However, when the cardinality $M$ of the  source images is not fixed a priori, as in our proposed task, a ``plain''  U-Net architecture cannot be used, being the number of input neurons a-priori fixed.
For this reason, we propose to modify the U-Net generator introducing an {\em attention} mechanism.
Attention is widely used to represent a variable-length input into a deep network \cite{DBLP:journals/corr/BahdanauCB14,DBLP:journals/corr/WestonCB14,DBLP:conf/nips/VinyalsFJ15,DBLP:conf/nips/VinyalsBLKW16,Dynamic-Few-Shot,DBLP:journals/corr/VinyalsBK15} and, without loss of generality, it can be thought of as a mechanism in which  multiple-input representations are averaged (i.e., summed) using some saliency criterion  emphasizing the importance of specific representations with respect to the others. In this paper we propose to use  attention in order to let the generator decide which specific image locations  
of each source image are the most trustable and informative at different convolutional layer resolutions. Specifically, we keep the standard encoder-decoder general partition typical of the U-Net (see Sec.~\ref{method})
but we propose three novelties. %% \footnote{Our code will be made publicly available after the acceptance of this article.}
 First, we introduce an {\em attention-based decoder} ($A$)
which  fuses the feature representations of each source. Second, we  encode the target pose and each source image with an {\em encoder} ($E$)
 %Network specifically designed to work synergistically with our attention-based decoder.
which 
 processes each source image $x_i$ {\em independently} of the others and $E$ locally deforms each $x_i$ performing a target-pose driven geometric ``normalization'' of $x_i$. Once normalized, the source images can be compared to each other in $A$, assigning location and source-specific saliency weights which are used for fusion.
Finally, we use a {\em multi-source} adversarial loss ${\cal L}_{M-GAN}$ that employs a single conditional discriminator to handle any arbitrary number of source images.

%% More details on our method are provided in Sec.~\ref{method}, after a brief overview of the related literature in Sec.~\ref{Related}. Our experiments are reported in Sec.~\ref{Experiments} and we conclude in Sec.~\ref{Conclusions}.

\section{Related work}
\label{Related}

Most of the image generation approaches are based either on Variational Autoencoders (VAEs) \cite{kingma2013auto} or on
  Generative Adversarial Networks (GANs) \cite{goodfellow2014generative}.
%   framework
%is one of the main paradigms used to generate images and it is based on two
% competing networks, a generator ($G$) and a discriminator ($D$). 
 %$G$ aims at generating realistic data which can  
%``fool'' $D$, which in turn  learns how to distinguish real from fake images.
GANs have been extended to conditional GANs \cite{DBLP:journals/corr/abs-1708-05789}, where the image generation depends on some input variable. For instance, in \cite{pix2pix2016}, an input image $x$ is ``translated'' into a different representation $y$ using a U-Net  generator.

The  person generation task (Sec.~\ref{Introduction}) is a specific case of a conditioned generation  process, where the conditioning variables are the source and the target images. Most of the previous works use conditional GANs and a U-Net architecture. For instance,
 Ma et al. \cite{ma2017pose} propose a two-step training procedure: pose generation and texture refinement, both obtained using a U-Net architecture.
  Recently, this work has been extended  in \cite{ma2018disentangled} by learning disentangled representations of the pose, the foreground  and the background. 
Following \cite{ma2017pose}, several methods for pose-guided image generation have been recently proposed
\cite{LassnerPG17,ZhaoWCLF17,siarohin2018deformable,balakrishnansynthesizing,si2018multistage}.
All these approaches  are based on the U-Net. However, the original U-Net, having a fixed-number of input images,  cannot be directly used for the multi-source image generation as defined in Sec.~\ref{Introduction}.
 Siarohin et al.~\cite{siarohin2018deformable} modify the U-Net   using {\em deformable skip connections} which align the input image features with the target pose. In this work we use an {\em encoder} similar to their proposal  in order to align the source images with the target pose, but we introduce a {\em pose stream} which compares the similarity between the source and the target pose. Moreover, similarly to the aforementioned works, also \cite{siarohin2018deformable} is single-source and uses a ``standard'' U-Net {\em decoder} \cite{pix2pix2016}.

%% Balakrishnan et al. \cite{balakrishnansynthesizing} partition the human body into different parts and separately deform each of them. However, the model training is based on pairs of conditioning images with the same background  in order to perform baackground substraction.In \cite{esser2018variational} a VAE is used to represent the target pose and source image appearance with two separeted encoder networks. The two descriptors are 
%%   cobined and given to a decoder network that ouputs the generated image. 

Other works on image-generation rely on a strong  supervision during training or testing. For instance,
 Neverova et al. \cite{Neverova_2018_ECCV} use a  dense-pose estimator \cite{Guler2018DensePose}  trained using image-to-surface correspondences \cite{Guler2018DensePose}.
 Dong et al. \cite{dong2018soft} use an externally trained model for image segmentation in order to improve the generation process.
 Zanfir et al. \cite{zanfir2018human} estimate the human 3D-pose using meshes and identify the mesh regions that can be transferred directly from the input image mesh to the target mesh.
 However, these methods cannot be directly compared with most of the other works, including ours, which rely only on a sparse keypoint detection.
Hard data-collection constraints are  used also in
\cite{chan2018dance}, where  a person and a background specific model are learned for video generation. This approach requires that the target person moves for several minutes covering all the possible poses  and that a new model is trained specifically for each target person.
Similarly, Liu et al. \cite{liu2018neural} compute the 3D human model by combining several minutes of video.
 In contrast with these works, our approach is based on fusing only a few source images in random poses 
and in variable number, which we believe is important because it makes it possible
to exploit existing datasets where multiple images are available for the same person.
Moreover, our network does not need to be trained for each specific person.

Sun et al. \cite{sun2018multi} propose a  multi-source image generation approach whose goal is to generate a new image according to a target-camera position. Note that
this task is different from what we address in this paper (Sec.~\ref{Introduction}), since 
a human pose describes an {\em articulated} object by means of a set of joint locations, while a camera position describes a viewpoint change but does not deal with 
 source-to-target  object {\em deformations}. Specifically, Sun et al. \cite{sun2018multi}
represent the camera pose with either a discrete label (e.g., \emph{left}, \emph{right},etc.) or a 6DoF vector
 and then they generate a pixel-flow which estimates the ``movement'' of each source-image pixel. Multiple images are integrated using a Convolutional LSTM \cite{DBLP:journals/corr/ShiCWYWW15} and confidence maps. 
 %One problem of this approach is that the 
 %pixel-flow prediction may be unable to generate large-region displacement differences which are common %of articulated objects.
 Most of the reported results concern 3D synthetic (rigid)  objects, while a few real scenes are also used but only with a limited viewpoint change.

\section{Attention-based U-Net}
\label{method}

\begin{figure*}[t]\centering
  \subfloat[A schematic representation of the proposed attention decoder architecture]{\includegraphics[width=0.65\linewidth]{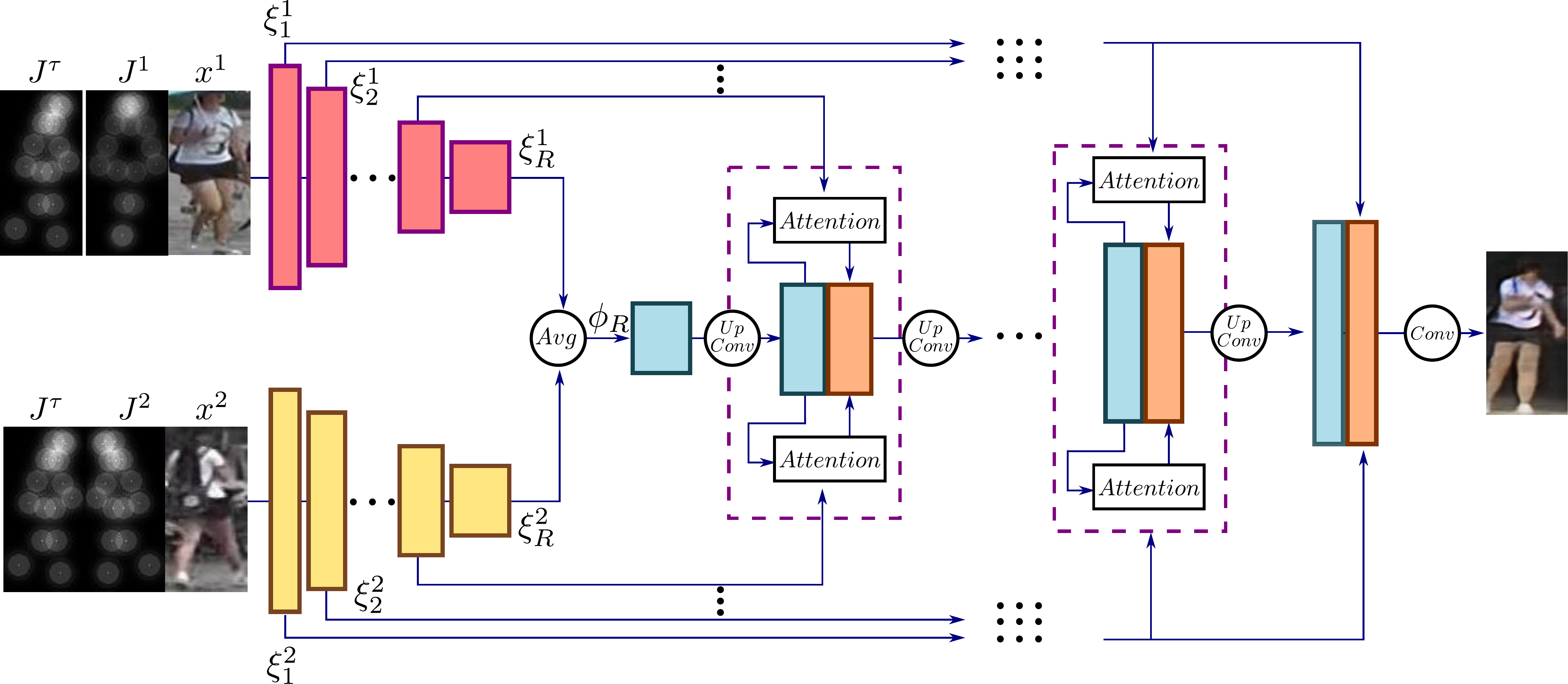}\label{sub:tab1}}
  \hspace{0.5cm}
\subfloat[Zoom on the attention module]{\includegraphics[width=0.25\linewidth]{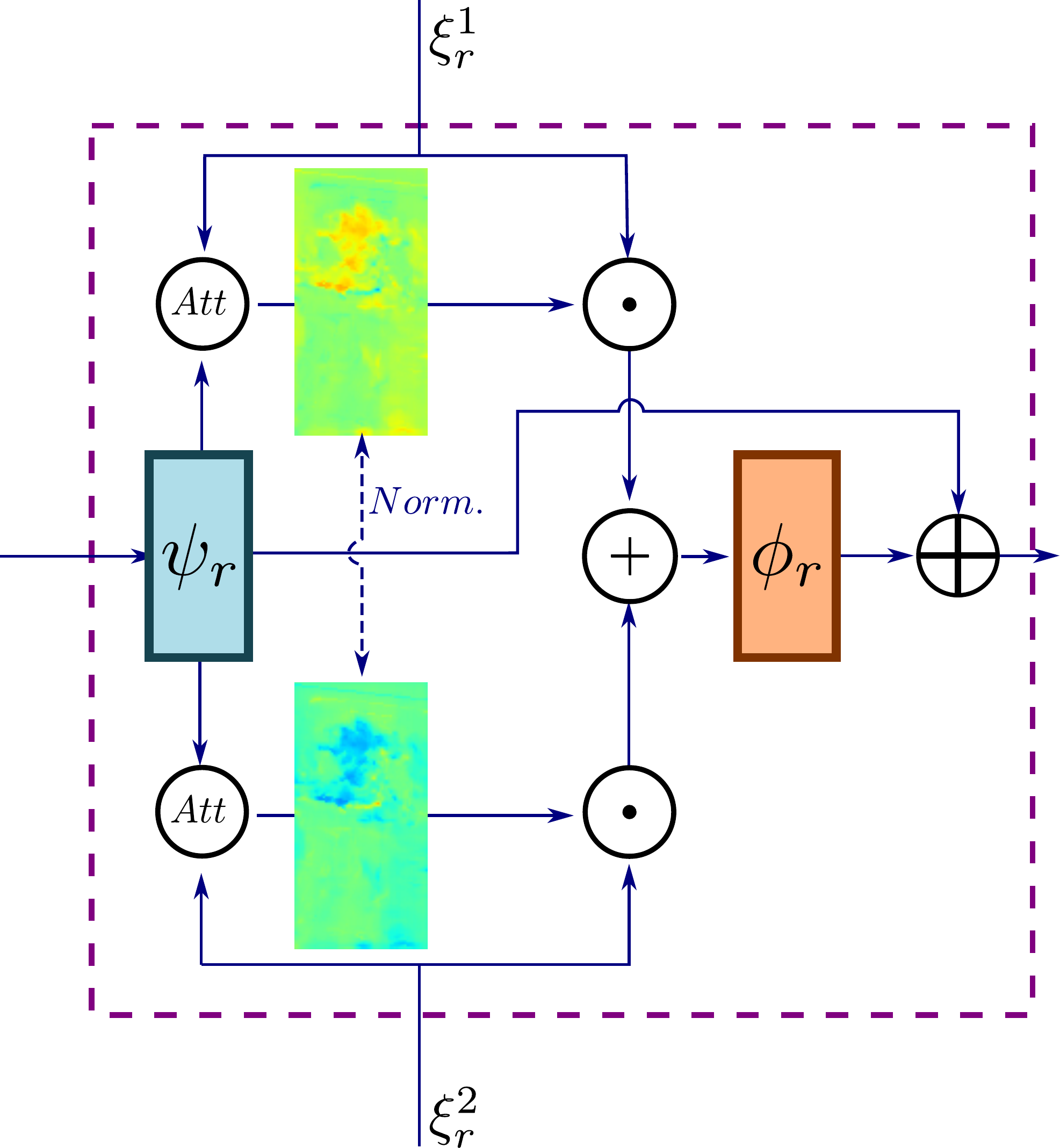}\label{sub:tab1}}
\caption{Illustration of the proposed Attention U-Net. For the sake of clarity, in this figure, we consider the case in which we use only two conditioning images ($M_n=2$). The colored rectangles represent the feature maps. The attention module (dashed purple rectangles) in the figure (a) are detailed in figure (b). The dashed double arrows denote normalization across attention maps, $\odot$ denotes the element-wise product and $\oplus$ denotes the concatenation along the channel axis.}
\vspace{-0.5cm}
\label{fig:att-decoder}
\end{figure*}

\subsection{Overview}
We first introduce some notation and provide a general overview of the proposed method.
Referring to the multi-source human image generation task defined in Sec.~\ref{Introduction}, we assume
 a training set $\mathcal{X} = \{\mathcal{X}_n\}_{n=1..N}$ is given, being each sample $\mathcal{X}_n=\left(\Xvect_n,\xtarget_n\right)$, where $\Xvect_n=\{\xsourcei_n\}_{i=1..M_n}$ is a set of $M_n$ source images of the same person sharing a common appearance and $\xtarget_n$ is the target image. Every sample image has the same size $H\times W$.
 Note that the source-set size $M_n$ is variable and depends on the person identity $n$. 
 Given an image $x$ depicting a person, we represent the body-pose as a set of 2D keypoints $P(x)= (\mathbf{p}_1, ... ,\mathbf{p}_K)$, where each $\mathbf{p}_k$ is the pixel location of a body joint in $x$. The body pose can be estimated from an image using an external HPE. The target pose is denoted by $\posetarget_n  = P(\xtarget_n)$. 
 
Our method is based on a conditional GAN approach, where the generator $G$ follows a general U-Net architecture \cite{pix2pix2016} composed of an encoder and a decoder.
A U-Net encoder is a sequence of convolutional and pooling layers, which progressively decrease the spatial resolution of the input representation. As a consequence, a specific activation in a given encoder layer has a receptive field progressively increasing with the layer depth, so gradually encoding ``contextual'' information. Vice versa, the decoder is composed of up-convolution layers, and, importantly, each decoder layer is connected to the corresponding layer in the encoder by means of {\em skip connections}, that concatenate the encoder-layer feature maps with the decoder-layer feature maps \cite{pix2pix2016}. Finally, Isola et al. \cite{pix2pix2016} use a conditional discriminator $D$ in order to discriminate between real and fake ``image transformations''.

We modify the aforementioned framework in three main aspects.
First, we use $M_n$ replicas of the same encoder  $E$ in order to encode the $M_n$ {\em geometrically normalized} source images together with the target pose. %% Its architecture based on \cite{siarohin2018deformable} in order to handles pose misalignments among the source images and with respect to the target pose by deforming the feature maps according to the pose differences.
Second, we propose an {\em attention-based decoder} $A$ that fuses the feature maps provided by the encoders.
Finally, we propose a {\em multi-source} adversarial loss ${\cal L}_{M-GAN}$.

%% $E$ (Sec.~\ref{sec:encoder}) is 
%%  ``single-source'' dependent, i.e., 
%% it  independently processes each 
%% $\xsourcei_n \in \Xvect_n$.
Fig.~\ref{fig:att-decoder} shows the
 architecture of $G$.
 Given a set $\Xvect_n$ of $M_n$ source images,  $E$ encodes each source image $\xsourcei_n \in \Xvect_n$ together with the target pose. Similarly to the standard U-Net, 
for a given source image $\xvect^{i}_n$,  
 each encoder outputs $R$ feature maps $\xivect_r^{i}\in\mathbb{R}^{H_r\times W_r\times C_r^E}, r \in [1..R]$ for  $R$ different-resolution blocks. 
%  In this work we  extend the U-Net using an attention mechanism whose goal is to combine  contextual %information produced by the  decoder upconvolutional layers and local and source-specific information %provided by each encoder tensor $\xivect_{r}^{i}$.
% \enver{i tried to modify your last sentence (you find it *commented* above) but it's still a bit hard %to understand... I think it's better % to completely drop it. an}
Each $\xivect_{r}^{i}$ is  aligned with the target pose (Sec~\ref{sec:encoder}).
This  alignment acts as a geometric ``normalization'' of each  $\xivect_{r}^{i}$ with respect to 
$\posetarget_n$ and makes it possible to compare  $\xivect_{r}^{i}$ with  $\xivect_{r}^{j}$ ($i \neq j$). 
 Finally, each  tensor $\xivect_{r}^{i}$ 
jointly represents pose and appearance information
at resolution $r$. 

\subsection{The Attention-based Decoder}
\label{sec:dec}

$A$ is composed of $R$ blocks. Similarly to the standard U-Net, the spatial resolution increases symmetrically with respect to the blocks in $E$. Therefore, to highlight this symmetry, the decoder blocks are indexed from R to 1. In the current $r$-th block, the image $\xgen$ which is going to be generated is represented by a tensor $\phivect_{r}$. This representation  is progressively refined in the subsequent blocks using an attention-based fusion of
$\{ \xivect_{r}^{i} \}_{i = 1,...,M_n}$. We call $\phivect_{r}$ the {\em latent representation} of $\xgen$ at resolution $r$, and $\phivect_{r}$ is recursively defined starting from $r= R$ till $r=1$ as follows: 

The initial latent representation $\phivect_R$ is obtained by averaging  
 the output tensors of the {\em last}  
 layer of $E$ (Fig.~\ref{fig:att-decoder}):
\begin{equation}
\label{eq:init-ave}
 \phivect_R=\frac{1}{M_n}\sum_{i=1}^{M_n} \xivect_R^{i}
\end{equation}
Note that each spatial position  in $\phivect_R$ corresponds to  a large receptive field in the original image resolution which, if $R$ is sufficiently large, may include  the whole initial image.
As a consequence, we can think of  $\phivect_R$ as encoding general contextual  information on  
$(\Xvect_n,\posetarget_n)$. 
 
For each subsequent block $r \in [R - 1, ..., 1]$, $\phivect_r$ is computed as follows.
Given $\phivect_{r+1}\in \mathbb{R}^{H_{r+1}\times W_{r+1}\times C_{r+1}^E}$,
we first perform a $2\times 2$ up-sampling 
on $\phivect_{r+1}$ followed by a convolution layer in order to obtain a tensor $\psivect_{r}\in \mathbb{R}^{H_{r}\times W_{r}\times C_{r}^D}$. 
$\psivect_{r}$ is then fed to an attention mechanism in order to estimate how the different 
tensors $\xivect_{r}^{i}$ should be fused into a single final tensor $F_{r}$:
\begin{equation}
\label{eq:phiAtt}
F_{r} = \sum_{i=1}^{M_n} Att(\psivect_{r},\xivect_{r}^{i})\odot\xivect_{r}^{i}, 
\end{equation}
\noindent
where $\odot$ denotes the element-wise product and $Att(\cdot,\cdot)\in [0,1]^{H_{r}\times W_{r}\times C_{r}^E}$ is the proposed attention module. 

In order to reduce the number of weights involved in computing 
Eq.~\eqref{eq:phiAtt}, we 
factorize 
$Att(\psivect_{r},\xivect_{r}^{i})$ using a spatial-attention 
$g(\psivect_{r},\xivect_{r}^{i}) \in [0,1]^{H_{r} \times W_{r}}$ (which is channel independent) and a channel-attention vector
$f(\psivect_{r},\xivect_{r}^{i}) \in [0,1]^{C_{r}^E}$ (which is spatial independent).
Specifically, at each spatial coordinate $(h,w)$, $g()$  compares the current latent representation 
$\psivect_{r}[h,w] \in \mathbb{R}^{C_{r}^D}$ with $\xivect_{r}^{i}[h,w] \in \mathbb{R}^{C_{r}^E}$ and assigns a saliency weight to 
$\xivect_{r}^{i}[h,w]$ which  represents how significant/trustable is $\xivect_{r}^{i}[h,w]$ with respect to $\psivect_{r}[h,w]$. The function
$g()$ is implemented by taking the concatenation of $\psivect_{r}$ and $\xivect_{r}^{i}$ as input and then using a
 $1 \times 1 \times (C_{r}^D + C_{r}^E)$ convolution layer. 
Similarly,  $f()$ 
is implemented by means of  global-average-pooling  on  the concatenation of $\psivect_{r}$ and $\xivect_{r}^{i}$
followed by  two fully-connected layers. We employ sigmoid activations on both $g$ and $f$.
Combining together $g()$ and $f()$, we obtain:
\begin{equation}
  A_{r}^{i}[h,w,c] =  g(\psivect_{r},\xivect_{r}^{i})[h,w] \cdot 
  f(\psivect_{r},\xivect_{r}^{i})[c].
  \label{eq:att}
\end{equation}
Importantly, $A_{r}^{i}$ is not spatially or channel normalized. This because  a normalization would enforce that, overall, each source image is used in the same proportion. Conversely, without normalization, given, for instance, a non-informative source $\xsourcei_n$ (e.g., $\xsourcei_n$ completely black), the attention module can correspondingly produce
 a null saliency tensor $A_{r}^{i}$. Nevertheless, the final attention tensor
  $Att()$ in  Eq.~\eqref{eq:phiAtt} is normalized in order to assign a {\em relative} importance to each source:
  \begin{equation}
 \label{eq.ineter-source-normalization}
  Att(\psivect_{r},\xivect_{r}^{i})[h,w,c] =  \frac{ A_{r}^{i}[h,w,c] }{ \sum_{j=1}^{M_n} A_{r}^{j}[h,w,c]}.
\end{equation}
Finally, the new latent representation at resolution $r$ is obtained by concatenating $\psivect_{r}$ with 
$F_{r}$:
\begin{equation}
\phivect_{r} = \psivect_{r} \oplus~ F_{r},
\end{equation}
\noindent
where $\oplus$ is the  tensor concatenation along the channel axis.

\subsection{The Pose-based Encoder}
\label{sec:encoder}

Rather than using a generic convolutional encoder as in \cite{pix2pix2016}, we use a task-specific encoder specifically designed
to work synergistically with our proposed attention model.
Our pose-based encoder $E$ is similar to the encoder proposed in \cite{siarohin2018deformable}
but it also contains a dedicated stream which is used to compare each other the source and the target pose. 
\begin{figure}[h]\centering
\includegraphics[width=0.8\linewidth]{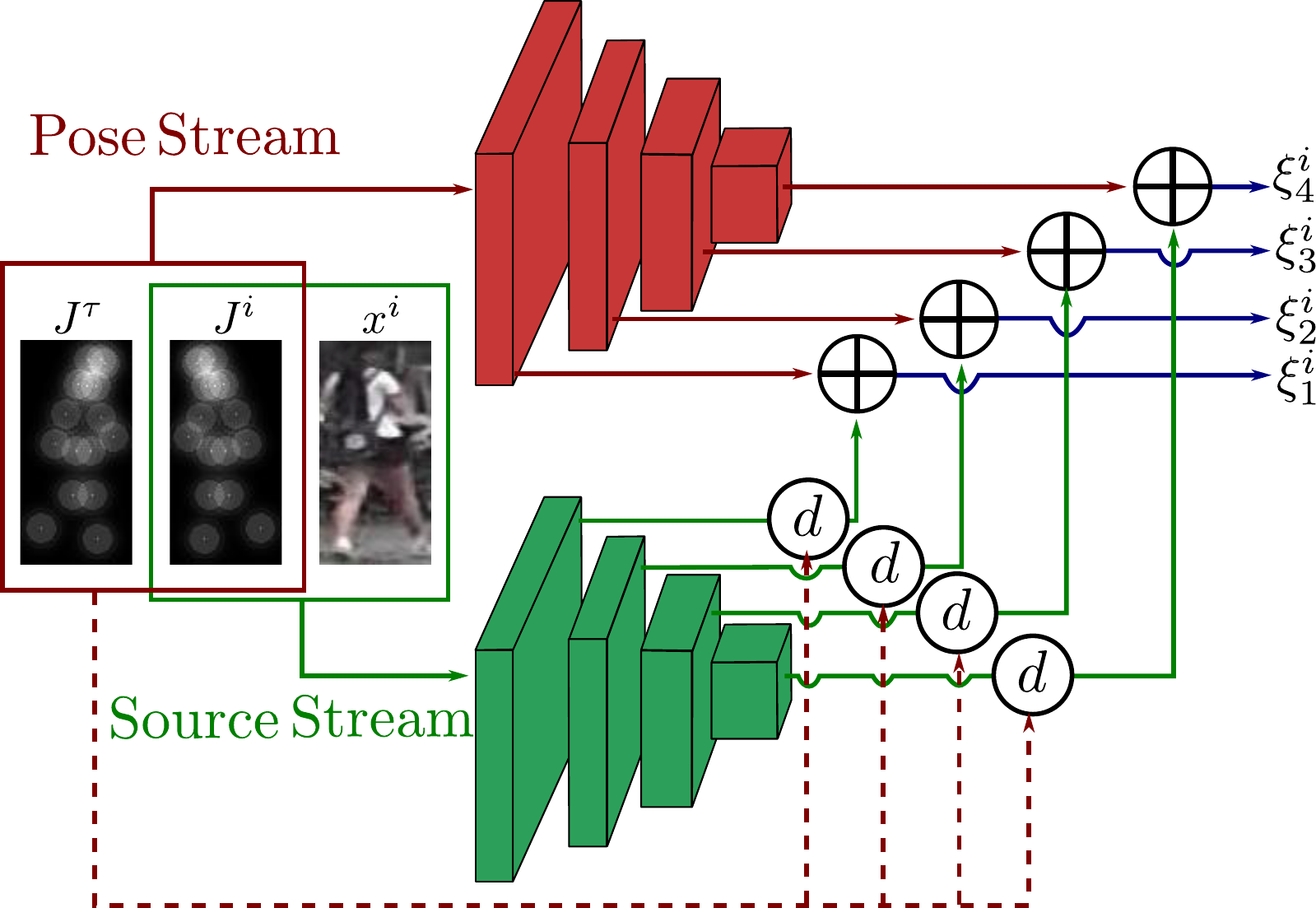}
\caption{The Pose-based encoder. For simplicity, we show only 4  blocks ($R=4$). Each parallelepiped  represents the feature maps obtained after convolution and max-pooling. The $d$ circles denote deformations.}
\vspace{-0.4cm}
\label{fig:encoder}
\end{figure}
In more detail, $E$ is composed of two streams (see Fig.~\ref{fig:encoder}). The first stream, referred to as {\em pose stream}, is used to represent pose information and to compare each other the target pose with the pose of the person in the source image. Specifically,
the target pose $p^\tau$ is represented using a tensor $J^\tau$ composed of $K$ heatmaps $J^{\tau,k} \in[0,1]^{H\times W}$. For each joint $\mathbf{p}^\tau_k \in p^\tau$, a heatmap $J^{\tau,k}$ is computed using a Gaussian kernel centered in $\mathbf{p}_k$ \cite{siarohin2018deformable}.
Similarly, given $\xsourcei_n \in \Xvect_n$, we extract the pose $P(\xsourcei_n)$ using \cite{Cao} and we describe it using a tensor $J^{i}_n$.
The tensors $J^{\tau}_n$ and $J^{i}_n$ are concatenated and input to the pose stream, which is composed of  a sequence of convolutional and pooling layers.
The purpose of the pose stream is twofold. First, it provides the target pose to the decoder. Second, it encodes the similarity 
between the $i$-th source pose and the target pose. This similarity is of a crucial importance for our attention mechanism to work
(Sec.~\ref{sec:dec})
 since a source image with a pose similar to the target pose is likely more trustable in order to transfer appearance information to the final generated image. For instance, a leg in $\xsourcei_n$ with a pose closer to $\posetarget_n$ than the corresponding leg in  $x^j_n$, should be most likely preferred for encoding the leg appearance.

 The second stream, called \emph{source stream}, takes as input the concatenation of the RGB image $\xsourcei_n$ and its pose representation 
$J^{i}_n$. $J^{i}_n$ is provided as input to the source stream in order to guide the source-stream convolutional layers in extracting relevant information which may depend on the joint locations.
The output of each convolutional layer of the source stream is a  tensor (green blocks in Fig.~\ref{fig:encoder}). This tensor 
is then {\em deformed} according to the difference between $P(\xsourcei_n)$ and  $\posetarget_n$
(the $d$ circles  in Fig.~\ref{fig:encoder}). 
 Specifically, we use body part-based affine deformations as in \cite{siarohin2018deformable} 
to locally deform the  source-stream feature maps at each given layer  and then concatenate the obtained tensor with  the corresponding-layer pose-stream tensor. In this way we  get a final tensor 
$\xivect_r^{i}$ for each of the $R$ different  layers in $E$ ($1 \leq r \leq R$). Each $\xivect_r^{i}$ is a representation of $(P(\xsourcei_n),\xsourcei_n)$ {\em aligned} with 
$\posetarget_n$ and it is obtained {\em independently} of $x^j_n \in \Xvect_n, j \neq i$.

 Given a set $\Xvect_n$ of $M_n$ source images, we apply $M_n$ replicas of the $E$ encoder to each $\xsourcei_n \in \Xvect_n$  producing the  set of 
 output tensors $\mathcal{E}_n = \{\xivect_r^{i} \}_{i = 1,...,M_n, r = 1,...R}$ that are input to the decoder described in Sec.\ref{sec:dec}.%%  Each 
%% $\xivect_r^{i}\in\mathbb{R}^{H_r\times W_r\times C_r^E}$ denotes the output of the $r^{th}$ block of $E$ (Fig.~\ref{fig:encoder}) with respect to the source image $\xsourcei_n$.

\subsection{Training}
\label{sec:training}

We  train the whole network in an end-to-end fashion combining a reconstruction loss with an adversarial loss.
For the reconstruction loss, we use the  {\em nearest-neighbour} loss $\mathcal{L}_{NN}(G)$ introduced in \cite{siarohin2018deformable} which exploits the convolutional maps of an external network (VGG-19 \cite{DBLP:journals/corr/SimonyanZ14a}, trained on ImageNet \cite{deng2009imagenet}) at the original image resolution in order to compare each  location of the generated image $\xgen$ with a local neighbourhood of the ground-truth 
image $\xtarget$.
This reconstruction loss is more robust to small spatial misalignments between $\xgen$ and $\xtarget$ than other common losses as the $L_1$ loss.

On the other hand, in our multi-source problem, the employed adversarial loss has to handle a varying number of sources.
We use a single-source discriminator conditioned on only one source image $\xsourcei_n$ \cite{pix2pix2016} %% and we employ this disciminator on every pair composed of a source image and the generated image. 
%% Our proposed 
%% {\em multi-source} 
%%  adversarial loss ${\cal L}_{M-GAN}$ is defined as follows.
More precisely, we use $M_n$ discriminators $D$ that share their parameters and independently process each $\xsourcei_n$. Each $D$ 
 takes as input the concatenation of four tensors: $x, J^\tau_n, \xsourcei_n, J^i_n$, where $x$ is either the ground truth real image $\xtarget_n$ or the generated image $\xgen$. Differently from other multi-source losses~\cite{yildirim2018disentangling,azadi2018multi,park2018mcgan}, we employ a conditional discriminator in order to exploit the information contained in the source image and the pose heatmaps. The GAN loss for the $i^{th}$ source image is defined as:
\begin{equation}
\label{eq.GAN-loss-i}
\begin{array}{cc}
{\cal L}_{GAN}^i(G,D)= &
\hspace{-7pt} \mathbb{E}_{(\xsourcei_n,\xtarget_n) \in {\cal X}} [\log D(\xtarget_n, J^\tau_n, \xsourcei_n, J^i_n)] + \\
 & \hspace{-40pt} \mathbb{E}_{(\xsourcei_n,\xtarget_n) \in {\cal X}, z \in {\cal Z}} [\log ( 1 - D(\xgen, J^\tau_n, \xsourcei_n, J^i_n) )],
\end{array}
\end{equation}
\noindent
where $\xgen = G(z, \Xvect_n, \posetarget_n)$ and, with a slight abuse of notation, $\mathbb{E}_{(\xsourcei_n,\xtarget_n) \in {\cal X}} [\cdot]$ means the expectation computed over pairs of single-source and target image extracted at random from the training set ${\cal X}$. %Differently from other multi-source losses~\cite{yildirim2018disentangling,azadi2018multi,park2018mcgan}, we use in \eqref{eq.GAN-loss-i} a conditional discriminator.
Using Eq.~\eqref{eq.GAN-loss-i}, the {\em multi-source} 
 adversarial loss (${\cal L}_{M-GAN}$) is defined as:
\begin{equation}
\label{eq.GAN-loss}
{\cal L}_{M-GAN}(G,D)=\min_G \max_D  \sum_{i=1}^{M_n}{\cal L}_{GAN}^i(G,D). 
\end{equation}

\noindent
Putting all together, the final training loss is given by:
\begin{equation}
\label{eq.objective}
G^* = \arg \min_G \max_D {\cal L}_{M-GAN}(G,D) + \lambda {\cal L}_{NN}(G),
\end{equation}
\noindent
where the $\lambda$ weight is  set to $ 0.01$ in all our experiments.

\section{Experiments}
\label{Experiments}

\begin{table*}[t]
\centering
\resizebox{0.95\textwidth}{!}{\begin{tabular}{l|c|cccc|cc}
  \toprule
  &&\multicolumn{4}{c|}{Market-1501}&\multicolumn{2}{c}{DeepFashion}\\
 Model&M &\emph{SSIM} & \emph{IS}&\emph{mask-SSIM} & \emph{mask-IS} & \emph{SSIM} & \emph{IS} \\
\midrule
Ma et al. \cite{ma2017pose} &1 &\bf$0.253$ & $3.460$ & $0.792$ & $3.435$&$0.762$ & $3.090$  \\
Ma et al. \cite{ma2018disentangled}&1  &\bf$0.099$ & $3.483$ & $0.614$ & $3.491$&$0.614$ & $3.228$   \\
Esser et al. \cite{esser2018variational}&1  &$\bf0.353$ & $3.214$ & $0.787$ & $3.249$  &$\bf0.786$ & $3.087$   \\
%% Esser et al. \cite{esser2018variational}&1  &$\bf0.353\pm0.10$ & $3.214\pm0.12$ & $0.787$ & $3.249$  &$\bf0.786\pm0.07$ & $3.087\pm0.24$   \\

Siarohin et al. \cite{siarohin2018deformable}&1 &$0.290$ & $3.185$ & $0.805$ & $3.502$& $0.756$ & $\bf3.439$ \\
\midrule
\emph{Ours}&1 &$0.270\pm0.09$& $3.251\pm0.09$ & $0.771\pm0.07$ &$3.614\pm0.08$& $0.757\pm0.07$ & $\bf3.420\pm0.06$  \\
\midrule
\emph{Ours}&2  &$0.285\pm0.09$ & $3.474\pm0.09$ & $0.778\pm0.06$ & $3.634\pm0.08$  &$0.769\pm 0.07$ & $3.421\pm0.06$ \\
\emph{Ours}&3  &$0.291\pm0.06$ & $3.442\pm0.09$ & $0.783\pm0.06$ & $3.739\pm0.08$  &$0.774\pm 0.07$ & $3.400\pm0.03$ \\
\emph{Ours}&5  &$0.306\pm0.09$ & $3.444\pm0.05$ & $0.788\pm0.06$ & $\bf3.814\pm0.07$   &$0.774\pm 0.06$ & $3.416\pm0.06$ \\
\emph{Ours}&7  &$0.320\pm0.09$ & $\bf3.613\pm0.05$ & $0.801\pm0.06$ & $3.567\pm0.06$  & - & -  \\
\emph{Ours}&10 &$0.326\pm0.09$ & $3.442\pm0.07$ & $\bf0.806\pm0.06$ & $3.514\pm0.04$  & - & -  \\
%% \midrule
%% \emph{Real-Data}&- &$1.00$ & $3.86$ & $1.00$ & $3.36$  &$1.000$ & $3.898$  \\
\bottomrule
\end{tabular}}

\caption{Comparison with the state of the art on the Market-1501 and the DeepFashion datasets. %% $(*)$ These values have been computed 
%% using the code and the network weights released by Ma et al. \cite{ma2017pose} in order to generate new images.
\label{tab:result}
}\vspace{-0.4cm}
\end{table*}

In this section we evaluate our method  both qualitatively and quantitatively adopting  the evaluation protocol proposed by Ma et al. \cite{ma2017pose}.
%% \subsection{Network and Training details}
\label{sec:impDetails}
We train $G$ and $D$ for 60k iterations, using the Adam optimizer (learning rate: $2 * 10^{-4}$, $\beta_1 = 0.5$, $\beta_2 =0.999$). We use instance normalization \cite{DBLP:journals/corr/UlyanovVL16} as recommended in \cite{pix2pix2016}. 
The networks used for $E$ and $D$ have  the same convolutional-layer dimensions and normalization parameters used in \cite{siarohin2018deformable}.
Also the up-convolutional layers of $A$ have the same dimensions of the corresponding decoder used in \cite{siarohin2018deformable}. Finally, the number of the hidden-layer neurons used to implement $f()$ (Sec.~\ref{sec:dec}) is $\frac{C_r^D + C_r^E}{4}$.
For a fair comparison with single-source person generation methods \cite{ma2017pose,ma2018disentangled,esser2018variational,siarohin2018deformable}, 
 we adopt the HPE proposed in \cite{Cao}. 
 
Even if there is no constraint on the  cardinality of the source images $M_n$, in order to simplify the implementation, we train and test our networks using different steps, each step having $M_n$ fixed for all $\mathcal{X}_n$ in  $\mathcal{X}$.
Specifically, we initially train $E$, $A$ and $D$ with $M_n = 2$. Then, we fine-tune the model with the desired $M_n$ value, except for single-source experiments where $M_n = 1$ (see Sec. \ref{sec:ablation}).

%% In the following we denote with: 
%% (1) $C_m^s$  a convolution layer with ReLU activation with $m$ filters and stride $s$,
%% (2)  $CN_m^s$  the same layer as $C_m^s$ with instance normalization before ReLU and
%% (3) $CD_m^s$ the same as $CN_m^s$  with the addition of dropout at rate $50\%$.
%% Both the source stream and transformation stream decoders have the following architecture is given by:
%% $CN_{64}^1 - CN_{128}^2 - CN_{256}^2 - CN_{512}^2 - CN_{512}^2 - CN_{512}^2$.
%% \noindent
%% The decoder is composed of the following convolution layers. 
%% $CD_{512}^2 - CD_{512}^2 - CD_{512}^2 - CN_{256}^2 - CN_{128}^2 - C_{3}^1$.
%% \noindent
%% In the last decoder layer,  the ReLU activation is replaced with $tanh$.

%% In order to handle the higher resolution of the DeepFashion dataset, we add one additional convolution block ($CN_{512}^2$) to 
%%  the encoder (both streams) and to the decoder.

\subsection{Datasets}
\label{expe:datasets}

The person re-identification  Market-1501 dataset \cite{zheng2015scalable} is composed of 32,668 images of 1,501 different persons captured from 6 surveillance cameras. This dataset is challenging because of the high diversity in pose,  background, viewpoint and illumination, and because of the low-resolution images (128$\times$64). To train our model, we need tuples of images of the same person in different poses. As this dataset is relatively noisy, we follow the preprocessing described in \cite{siarohin2018deformable}. The images where no human body is detected using the HPE are removed. Other methods \cite{ma2017pose,ma2018disentangled,esser2018variational,siarohin2018deformable} generate all the possible pairs for each identity. However, in our approach, since we consider tuples of size $M+1$ ($M$ sources and 1 target image), considering all the possible tuples is computationally infeasible. In addition, Market-1501 suffers from a high person-identity imbalance and computing all the possible tuples, would exponentially increase this imbalance. Hence, we  generate tuples randomly in such a way that we obtain the same identity repartition than it is obtained when sampling all the possible pairs. In addition, this solution also allows for  a fair comparison with single-source  methods which sample based on pairs. Eventually, we get 263K tuples for training.
For testing, following  \cite{ma2017pose}, we  randomly select 12K tuples without person is in common between the training and the test split.
 
The DeepFashion dataset ({\em In-shop Clothes Retrieval Benchmark}) \cite{liu2016deepfashion} consists of 52,712 clothes images with a resolution of 256$\times$256 pixels. For each outfit, we dispose of about 5 images with different viewpoints  and poses. Thus, we only perform experiments using up to $M_n = 5$ sources.
Following the training/test split adopted in \cite{ma2017pose}, we create tuples of images following the same protocol as for the market-1501 dataset.
After removing the images where the HPE does not detect any human body, we finally 
 collect about 89K tuples for training and 12K tuples for testing.

\subsection{Metrics}
\label{sec:metrics}
Evaluation metrics in the context of generation tasks is a problem in itself. In our experiments we adopt the evaluation metrics proposed in \cite{ma2017pose} which is used by most of the single-source methods. Specifically, we use: Structural Similarity (\emph{SSIM}) \cite{wang2004image}, Inception Score (\emph{IS}) \cite{salimans2016improved} and their corresponding masked versions \emph{mask-SSIM} and \emph{mask-IS} \cite{ma2017pose}. The masked versions of the metrics  are obtained by masking-out the image background. The motivation behind the use of masked metrics is that no background information is given to the network, and therefore, the network cannot guess the correct background of the target image.
For a fair comparison, we adopt the masks as defined in \cite{ma2017pose}.

It is worth noting that the SSIM-based metrics  compare the generated image with the ground-truth. Thus, they measure how well the model transfers the appearance of the person from the source image. Conversely, IS-based metrics evaluate the distribution of generated images, jointly assessing the degree of realism and  diversity of the generated outcomes, but do not take into account any similarity 
with the conditioning variables.
 These two metrics are each other complementary~\cite{blau2018perception} and should be interpreted jointly.

%% Finally, in our tables we also include  the value of each metrics computed using the {\em real} images of the test set. Since these  values are computed on real data, they can be considered as a sort of an upper-bound to the results a generator can obtain. However, these  values are not actual upper bounds in the strict sense: for instance the DS metrics on the real datasets is not 1 because of SSD failures. 

\subsection{Comparison with previous work}
\label{Comparison}

\begin{figure*}[h]
  \centering
  \setlength\tabcolsep{1.0pt}
  \resizebox{0.95\textwidth}{!}{\begin{tabular}{c|c|ccc|ccc|c}
    %% &&\multicolumn{4}{c|}{Market-1501}&\multicolumn{2}{c}{DeepFashion}\\
    $x^i,i\in[1..5]$ & $x_\tau$ & \cite{siarohin2018deformable}&  \cite{esser2018variational} & \cite{ma2017pose} &\multicolumn{3}{c|}{Ours }&\small Attention Saliency\\
    &&\multicolumn{3}{c|}{$M_n=1$} &$M_n=1$ &$M_n=3$ &$M_n=5$ &$M_n=5$ \\
%% \includegraphics[height=0.24\columnwidth]{figures/market/0from.jpg}
%% &\includegraphics[height=0.24\columnwidth]{figures/market/0to.jpg}
%% &\includegraphics[height=0.24\columnwidth]{figures/market/0fm.jpg}
%% &\includegraphics[height=0.24\columnwidth]{figures/market/0vunet.png}
%% &\includegraphics[height=0.24\columnwidth]{figures/market/0ma.jpg}
    %% \\

\includegraphics[height=0.24\columnwidth]{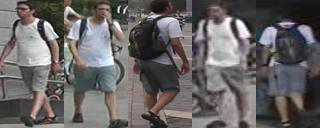}
&\includegraphics[height=0.24\columnwidth]{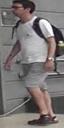}
&\includegraphics[height=0.24\columnwidth]{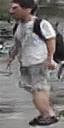}
&\includegraphics[height=0.24\columnwidth]{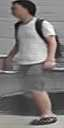}
&\includegraphics[height=0.24\columnwidth]{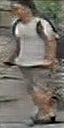}
&\includegraphics[height=0.24\columnwidth]{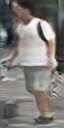}
&\includegraphics[height=0.24\columnwidth]{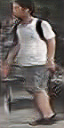}
&\includegraphics[height=0.24\columnwidth]{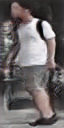}
&\includegraphics[height=0.24\columnwidth]{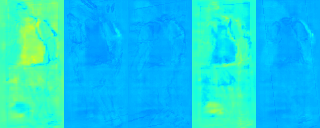}
\\
%% \includegraphics[height=0.24\columnwidth]{figures/market/2from.jpg}
%% &\includegraphics[height=0.24\columnwidth]{figures/market/2to.jpg}
%% &\includegraphics[height=0.24\columnwidth]{figures/market/2fm.jpg}
%% &\includegraphics[height=0.24\columnwidth]{figures/market/2vunet.png}
%% &\includegraphics[height=0.24\columnwidth]{figures/market/2ma.jpg}
%% \\
\includegraphics[height=0.24\columnwidth]{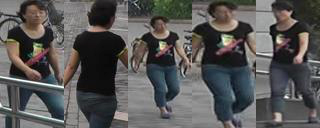}
&\includegraphics[height=0.24\columnwidth]{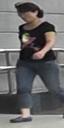}
&\includegraphics[height=0.24\columnwidth]{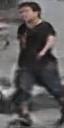}
&\includegraphics[height=0.24\columnwidth]{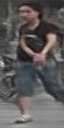}
&\includegraphics[height=0.24\columnwidth]{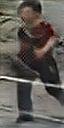}
&\includegraphics[height=0.24\columnwidth]{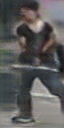}
&\includegraphics[height=0.24\columnwidth]{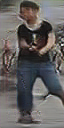}
&\includegraphics[height=0.24\columnwidth]{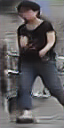}
&\includegraphics[height=0.24\columnwidth]{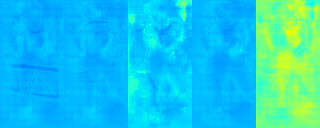}
\\
%% \includegraphics[height=0.24\columnwidth]{figures/market/4from.jpg}
%% &\includegraphics[height=0.24\columnwidth]{figures/market/4to.jpg}
%% &\includegraphics[height=0.24\columnwidth]{figures/market/4fm.jpg}
%% &\includegraphics[height=0.24\columnwidth]{figures/market/vunet.png}
%% &\includegraphics[height=0.24\columnwidth]{figures/market/4ma.jpg}
%% \\
%% \includegraphics[height=0.24\columnwidth]{figures/market/5from.jpg}
%% &\includegraphics[height=0.24\columnwidth]{figures/market/5to.jpg}
%% &\includegraphics[height=0.24\columnwidth]{figures/market/5fm.jpg}
%% &\includegraphics[height=0.24\columnwidth]{figures/market/vunet.png}
%% &\includegraphics[height=0.24\columnwidth]{figures/market/5ma.jpg}
%% \\
%% \includegraphics[height=0.24\columnwidth]{figures/market/6from.jpg}
%% &\includegraphics[height=0.24\columnwidth]{figures/market/6to.jpg}
%% &\includegraphics[height=0.24\columnwidth]{figures/market/6fm.jpg}
%% &\includegraphics[height=0.24\columnwidth]{figures/market/vunet.png}
%% &\includegraphics[height=0.24\columnwidth]{figures/market/6ma.jpg}
%% \\
\includegraphics[height=0.24\columnwidth]{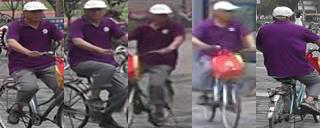}
&\includegraphics[height=0.24\columnwidth]{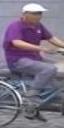}
&\includegraphics[height=0.24\columnwidth]{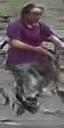}
&\includegraphics[height=0.24\columnwidth]{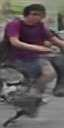}
&\includegraphics[height=0.24\columnwidth]{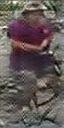}
&\includegraphics[height=0.24\columnwidth]{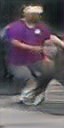}
&\includegraphics[height=0.24\columnwidth]{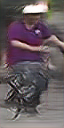}
&\includegraphics[height=0.24\columnwidth]{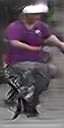}
&\includegraphics[height=0.24\columnwidth]{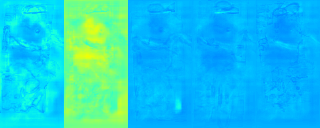}
\\
%% \includegraphics[height=0.24\columnwidth]{figures/market/8from.jpg}
%% &\includegraphics[height=0.24\columnwidth]{figures/market/8to.jpg}
%% &\includegraphics[height=0.24\columnwidth]{figures/market/8fm.jpg}
%% &\includegraphics[height=0.24\columnwidth]{figures/market/8vunetori.png}
%% &\includegraphics[height=0.24\columnwidth]{figures/market/8ma.jpg}
%% \\
\includegraphics[height=0.24\columnwidth]{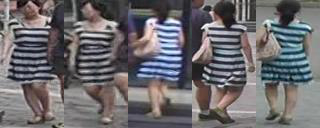}
&\includegraphics[height=0.24\columnwidth]{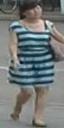}
&\includegraphics[height=0.24\columnwidth]{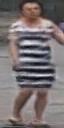}
&\includegraphics[height=0.24\columnwidth]{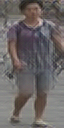}
&\includegraphics[height=0.24\columnwidth]{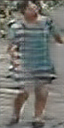}
&\includegraphics[height=0.24\columnwidth]{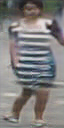}
&\includegraphics[height=0.24\columnwidth]{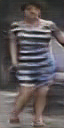}
&\includegraphics[height=0.24\columnwidth]{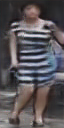}
&\includegraphics[height=0.24\columnwidth]{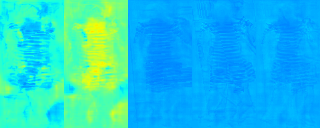}
\\
\includegraphics[height=0.24\columnwidth]{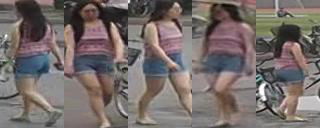}
&\includegraphics[height=0.24\columnwidth]{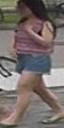}
&\includegraphics[height=0.24\columnwidth]{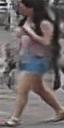}
&\includegraphics[height=0.24\columnwidth]{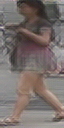}
&\includegraphics[height=0.24\columnwidth]{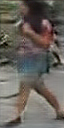}
&\includegraphics[height=0.24\columnwidth]{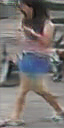}
&\includegraphics[height=0.24\columnwidth]{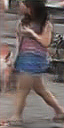}
&\includegraphics[height=0.24\columnwidth]{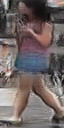}
&\includegraphics[height=0.24\columnwidth]{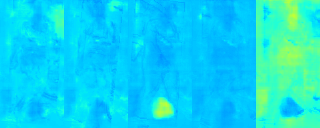}

  \end{tabular}}
  \vspace{-0.4cm}
  \caption{A qualitative comparison on the Market-1501 dataset.  The first column shows the source  images. Note that \cite{siarohin2018deformable,ma2017pose,esser2018variational} use only the leftmost source image. The target poses are given by the ground truth images in  column 2. In column 4, we show the results obtain by our model while increasing the number of source images. The source from the first column are added while increasing $M_n$ from left to right. In the last column we show the saliency maps predicted by our model when using all the five source images. These maps are shown in the same order than the source images $x^i$.}
\label{fig:comparison-Market-2}
\end{figure*}

{\bf Quantitative comparison.}
In Tab.~\ref{tab:result} we show a
 quantitative comparison with state-of-the-art single-source methods. 
 Note that, except from \cite{ma2018disentangled}, none of the compared methods, including ours, is conditioned on background information. On the other hand, the mask-based metrics focus on only the region of interest (i.e., the foreground person) and they are not biased by the randomly generated background. For these reasons, we believe the mask-based metrics are the most informative ones. However,
on the DeepFashion dataset, following  \cite{ma2018disentangled}, 
 we do not report the masked values being the background uniform in most of the images.
 On both datasets, we observe that the SSIM and masked-SSIM increase when we input more images to our model. This confirms the idea that multi-source image generation is an effective direction to improve the generation quality. Furthermore, it illustrates that the proposed model is able to combine the information provided by the different source images. Interestingly, our method reaches high SSIM scores while keeping high IS values, thus showing that it is able to transfer better the appearance without loosing image quality and diversity. 
 
 Concerning the comparison with the state of the art, our method reports the highest performance according to both the mask-SSIM and the mask-IS metrics on the Market-1501 dataset when we use 10 source images. When we employ fewer images, only Siarohin et al \cite{siarohin2018deformable} obtain better masked-SSIM but at the cost of a significantly lower IS.
 Similarly, we observe that  \cite{esser2018variational} achieves a really high SSIM score, but again at the cost of a drastically lower IS, meaning that we can generate more diverse and higher quality images. Moreover, we notice that \cite{esser2018variational} obtains a lower masked-SSIM. This seems to indicate that their high SSIM score is mostly due to a better background generation.
Similar conclusions can be drawn for the DeepFashion dataset. We obtain the best IS and rank second in SSIM. Only \cite{esser2018variational} outperforms our model in terms of SSIM at the cost of a much lower IS value. The gain in performance seems smaller than on the market-1501 dataset. This is probably due to the lower  pose diversity  of the DeepFashion dataset.

{\bf Qualitative comparison.}
Fig.~\ref{fig:comparison-Market-2} shows some images obtained using the Market-1501 dataset. We compare our results with the images generated by three methods for which the code is publicly available \cite{esser2018variational,ma2017pose,siarohin2018deformable}. The source images are shown in the first column. Note that the single-source methods use only the leftmost image. The target pose is extracted from the ground-truth target image. We display the generated images varying $M_n \in \{ 1, 3,  5 \}$. We also show the corresponding saliency tensors $A_r^i$ (see Sec.~\ref{sec:dec}) at the highest resolution $r= 1$. Specifically, we use $M_n = 5$ and, at each $(h,w)$ location in $A_r^i$, we average the values  over the channel axis ($c$) using a color scale from dark blue (0 values) to orange (1 values).

The qualitative results confirm  the quantitative evaluation  since we clearly  obtain better images when we increase the number of source images. The images become sharper and with more details and contain less artifacts. By looking at the saliency maps, we observe that our model uses mostly the source images in wich the human pose is similar to the target pose. For instance in row 1 and 4, the model has high attention values for the two frontal images but very low values for the back view images. Interestingly, in row 1, among the two source images with a pose similar to the target pose, the  saliency values are lower for the more blurry image. This illustrates that, between two images with similar poses, our attention model favours the image with the highest quality.
Concerning the comparison with the state of the art, we observe that our model better preserves the details of the source images. In general, we obtain higher-quality details and less artefacts. For instance, in row 3, the three other methods do not generate the white hat nor the small logo of the shirt. In particular, the V-UNet architecture proposed in  \cite{esser2018variational} generates realistic images but with less accurate details. This can be easily observed in the last two rows where the colors of the clothes are wrongly generated.

\subsection{Ablation study and qualitative analysis}
\label{sec:ablation}

\label{Ablation}
In this section we present  an ablation study to clarify the impact of each part of our proposal on the final performance. We first describe the compared methods, obtained by ``amputating'' important parts of the full-pipeline presented in Sec.~\ref{method}. The discriminator architecture is the same for all the methods. 
\begin{itemize}[noitemsep,topsep=0pt]\setlength\itemsep{0em}
\item \emph{Avg No-d}: In this baseline version of our method we use the encoder described in 
Sec.~\ref{sec:encoder} {\em without} the deformation-based alignment of the features with the target pose. For the decoder, we use a standard U-Net decoder without attention module. More precisely, the tensors provided by the skip connections of each encoder are simply averaged and concatenated with the decoder tensors as in the original U-Net.  In other words, Eq.~\eqref{eq:phiAtt} is replaced by the average over each convolution layer of the decoder, similarly to \eqref{eq:init-ave}.
\item \emph{Avg}: We use the encoder  described in Sec.~\ref{sec:encoder} and the same decoder of \emph{Avg No-d}.
\item \emph{Att. 2D}: We use an attention model similar to the full model described in Sec. \ref{sec:dec}. However, in Eq.~\eqref{eq:att}, $f(\cdot,\cdot)[c]$ is not used.
% and it is replaced by a vector of $1$.
\item \emph{Full}: This is the full-pipeline as described in Sec.~\ref{method}.
\end{itemize}

\begin{table}[h]
\centering
\resizebox{0.99\columnwidth}{!}{\begin{tabular}{l|c|cccc|cc}
  \toprule
  &&\multicolumn{4}{c|}{Market-1501}&\multicolumn{2}{c}{DeepFashion}\\
\midrule
  Model &$M_n$&\emph{SSIM} & \emph{IS}&\emph{mask-SSIM} & \emph{mask-IS} &\emph{SSIM} & \emph{IS}\\
\midrule
\emph{Single source}&1&$0.27$& $3.251$ & $0.771$ &$3.614$ & $0.757$ & $3.420$  \\
\midrule  
\emph{Avg No-d}&2& $0.258$& $3.182$ & $0.766$& $3.658$   & $0.756$   & $3.274$  \\
\emph{Avg }&2&  $\bf0.294$ & $3.468$ &  $\bf0.779$&  $3.274 $     &$0.785$ & $3.321$ \\

\emph{Att. 2D}&2&$0.285$ &  $3.460$ & $0.777$ & $3.632$ &$\bf0.769$   & $3.375$  \\
\emph{Full }&2&$0.285$ & $\bf3.474$ & $0.778$ & $\bf3.634$  &$\bf0.769$ & $\bf3.421$ \\
\midrule
\emph{Avg}&5&  $0.299$ & $3.383$ &  $0.782$&   $3.751$    & $0.763$ & $\bf3.454$ \\
\emph{Att. 2D}&5&$\bf 0.308$ &  $3.159$ & $\bf0.792$ & $3.606$ &$0.773$   & $3.411$  \\
\emph{Full}&5&$0.306$ & $\bf3.444$ & $0.788$ & $\bf3.814$   &$\bf0.774$ & $3.416$ \\

%% \midrule
%% \emph{Real-Data}&$1.00$ & $3.86$ & $1.00$ & $3.36$ &$1.000$ & $3.898$ \\
\bottomrule 
  \end{tabular}}
\caption{Quantitative ablation study on the Market-1501 and the DeepFashion dataset.}
  \label{tab:ablation}
\end{table}  \vspace{-0.1cm}

\begin{figure}[h]
  \centering
  \setlength\tabcolsep{1.0pt}
  \resizebox{0.99\columnwidth}{!}{\begin{tabular}{c|c|ccc}
    %% &&\multicolumn{4}{c|}{Market-1501}&\multicolumn{2}{c}{DeepFashion}\\
    $x^i,i\in[1..2]$ & $x_\tau$ & Avg    &Full & \small Attention Saliency\\

\includegraphics[height=0.24\columnwidth]{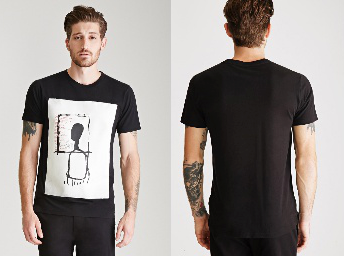}
&\includegraphics[height=0.24\columnwidth]{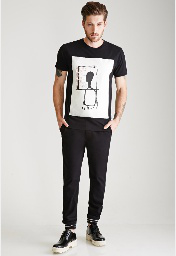}
&\includegraphics[height=0.24\columnwidth]{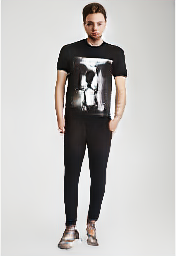}
&\includegraphics[height=0.24\columnwidth]{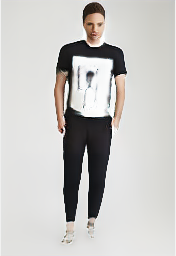}
&\includegraphics[height=0.24\columnwidth]{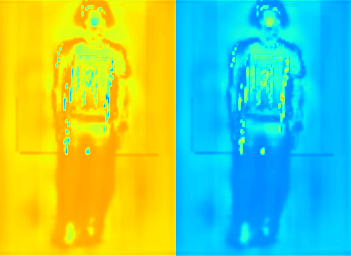}\\

\includegraphics[height=0.24\columnwidth]{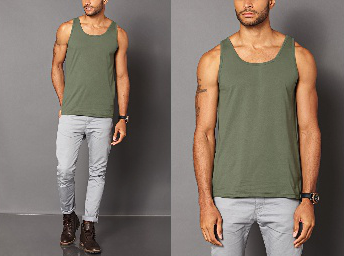}
&\includegraphics[height=0.24\columnwidth]{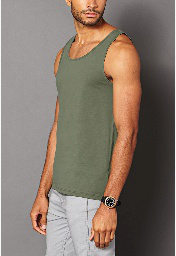}
&\includegraphics[height=0.24\columnwidth]{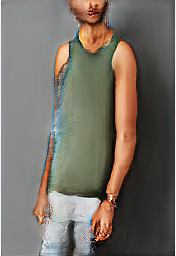}
&\includegraphics[height=0.24\columnwidth]{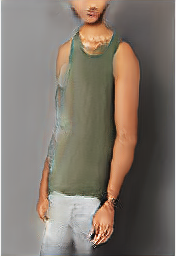}
&\includegraphics[height=0.24\columnwidth]{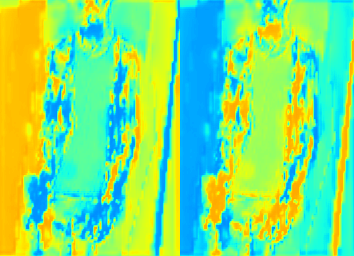}\\

%% \includegraphics[height=0.24\columnwidth]{figures/fashionpaper/input_fasionWOMENBlousesShirtsid0000023504_4full.png}
%% &\includegraphics[height=0.24\columnwidth]{figures/fashionpaper/GT_fasionWOMENBlousesShirtsid0000023504_4full.png}
%% &\includegraphics[height=0.24\columnwidth]{figures/fashionpaper/avg/fasionWOMENBlousesShirtsid0000023504_4full.png}
%% &\includegraphics[height=0.24\columnwidth]{figures/fashionpaper/full/fasionWOMENBlousesShirtsid0000023504_4full.png}
%% &\includegraphics[height=0.24\columnwidth]{figures/fashionpaper/att_fasionWOMENBlousesShirtsid0000023504_4full.png}\\

%% \includegraphics[height=0.24\columnwidth]{figures/fashionpaper/input_fasionWOMENBlousesShirtsid0000173903_2side.png}
%% &\includegraphics[height=0.24\columnwidth]{figures/fashionpaper/GT_fasionWOMENBlousesShirtsid0000173903_2side.png}
%% &\includegraphics[height=0.24\columnwidth]{figures/fashionpaper/avg/fasionWOMENBlousesShirtsid0000173903_2side.png}
%% &\includegraphics[height=0.24\columnwidth]{figures/fashionpaper/full/fasionWOMENBlousesShirtsid0000173903_2side.png}
%% &\includegraphics[height=0.24\columnwidth]{figures/fashionpaper/att_fasionWOMENBlousesShirtsid0000173903_2side.png}\\

\includegraphics[height=0.24\columnwidth]{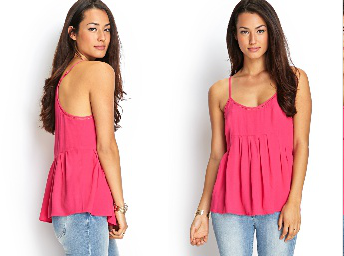}
&\includegraphics[height=0.24\columnwidth]{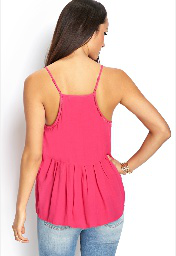}
&\includegraphics[height=0.24\columnwidth]{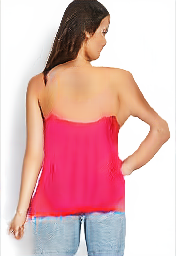}
&\includegraphics[height=0.24\columnwidth]{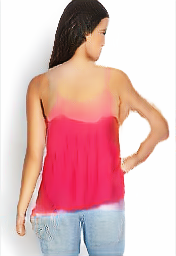}
&\includegraphics[height=0.24\columnwidth]{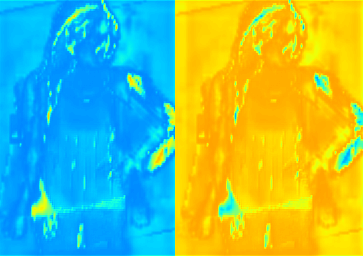}\\

%% \includegraphics[height=0.24\columnwidth]{figures/fashionpaper/input_fasionWOMENBlousesShirtsid0000191502_2side.png}
%% &\includegraphics[height=0.24\columnwidth]{figures/fashionpaper/GT_fasionWOMENBlousesShirtsid0000191502_2side.png}
%% &\includegraphics[height=0.24\columnwidth]{figures/fashionpaper/avg/fasionWOMENBlousesShirtsid0000191502_2side.png}
%% &\includegraphics[height=0.24\columnwidth]{figures/fashionpaper/full/fasionWOMENBlousesShirtsid0000191502_2side.png}
%% &\includegraphics[height=0.24\columnwidth]{figures/fashionpaper/att_fasionWOMENBlousesShirtsid0000191502_2side.png}\\

\end{tabular}}  \vspace{-0.4cm}
  \caption{A qualitative ablation study on the Deep-Fashion dataset. We compare  \emph{Avg}  with  \emph{Full}  using $M_n = 2$. The attention saliency are displayed in the same order than the source images $x^i$.}
\label{fig:ablation-deepFashion}
\vspace{-0.4cm}
\end{figure}  

\noindent
Tab.~\ref{tab:ablation} shows a quantitative evaluation. First, we notice that our method without  spatial deformation performs poorly on both datasets. This is  particularly evident with the \emph{SSIM}-based scores. This confirms the importance of source-target alignment before computing a position-dependent attention. Interestingly, when using only two source images, \emph{Avg}, \emph{Att. 2D} and \emph{Full} perform similarly to each other on the Market-1501 dataset. However, when we dispose of more source images we clearly observe the benefit of using our proposed attention approach.  \emph{Avg}  performs constantly worst than our \emph{Full} pipeline. The 2D attention model outputs images with higher \emph{SSIM}-based scores but with lower IS values. Concerning the DeepFashion dataset, our attention model performs that the simpler approach with 2 and 5 source images.

In Fig.~\ref{fig:ablation-deepFashion} we compare \emph{Avg}  with  \emph{Full}  using $M_n = 2$. 
The advantage of using \emph{Full} is
 is clearly illustrated by the fact that \emph{Avg} mostly performs an average of the front and back images. In the second row, \emph{Full} reduces the amount of artefacts. Interestingly, in the last row, \emph{Full} fails to generate correctly the new viewpoint but we see that it chooses to focus on the back view in order to generate  the collar.

\section{Conclusion}
\label{Conclusions}
In this work we introduced a generalization of the person-image generation problem. Specifically, a human image is generated conditioned on a target pose and  {\em a set} $\Xvect$ of source  images. This makes it possible to exploit multiple and possibly complementary images. We introduced an attention-based decoder which extends the U-Net architecture to a  multiple-input setting. Our attention mechanism selects relevant information from different sources and image regions. We experimentally validate our approach on two different datasets. We expect that the practical advantages of the multi-source approach, as demonstrated in this work, will attract the interest of the community.

\label{Conclusion}

\section*{Acknowledgements}

We thank the
NVIDIA Corporation for the donation of the GPUs used
in this work. This project has received funding from
the European Research Council (ERC) (Grant agreement
No.788793-BACKUP).

{\small
\bibliographystyle{ieee}
\bibliography{egbib}
}

\end{document}